\documentclass{bmvc2k}

\usepackage{array}
\usepackage{floatrow}

\title{The Role of Context Selection \\ in Object Detection}

\addauthor{Ruichi (Rich) Yu}{richyu@cs.umd.edu}{1}
\addauthor{Xi (Stephen) Chen }{chnxi@microsoft.com}{2}
\addauthor{Vlad I. Morariu}{morariu@umiacs.umd.edu}{1}
\addauthor{Larry S. Davis}{lsd@umiacs.umd.edu}{1}

\addinstitution{
 University of Maryland \\ College Park, MD. USA.
}
\addinstitution{
Microsoft Corporation\\
One Microsoft Way,\\
Redmond, WA. USA.
}

\runninghead{Yu \lowercase{et al.}}{The Role of Context Selection in Object Detection}

\begin{document}

\maketitle

\begin{abstract}
We investigate the reasons why context in object detection has limited utility by isolating and evaluating the predictive power of different context cues under ideal conditions in which context provided by an oracle. Based on this study, we propose a region-based context re-scoring method with dynamic context selection to remove noise and emphasize informative context. We introduce latent indicator variables to select (or ignore) potential contextual regions, and learn the selection strategy with latent-SVM. We conduct experiments to evaluate the performance of the proposed context selection method on the SUN RGB-D dataset. The method achieves a significant improvement in terms of mean average precision (mAP), compared with both appearance based detectors and a conventional context model without the selection scheme.\end{abstract}

\section{Introduction}
Context captures statistical and common sense properties of the real-world and plays a critical role in perceptual inference \cite{Mottaghi_2014_CVPR}. There are numerous studies that demonstrate the advantages of context in object recognition \cite{conf/cvpr/DivvalaHHEH09,Mottaghi_2014_CVPR,NIPS2003_2386,Galleguillos2008,chen2016object}. In contrast, other investigations have revealed situations in which context does not improve the performance of object detection \cite{Lin_2013_ICCV,DBLP:conf/cvpr/YaoFU12}, and sometimes introducing context even decreases performance \cite{DBLP:conf/cvpr/YaoFU12}. Additionally, driven by the development of deep CNNs \cite{NIPS2012_4824,Simonyan14c}, the performance of object detection has been dramatically boosted \cite{DBLP:journals/corr/GirshickDDM13,DBLP:journals/corr/Girshick15,NIPS2015_5638,DBLP:journals/corr/GuptaHM15}. While context has been incorporated into deep learning frameworks, the performance gain from context itself has not been significant \cite{crf,Attentive}. This leads to the question: how important is context in object detection when we have reasonably good detectors?

\begin{figure} 
  \centering 
  \subfigure[Detection Results]{ 
    \label{fig:subfig:a} 
    \includegraphics[width=0.30\linewidth]{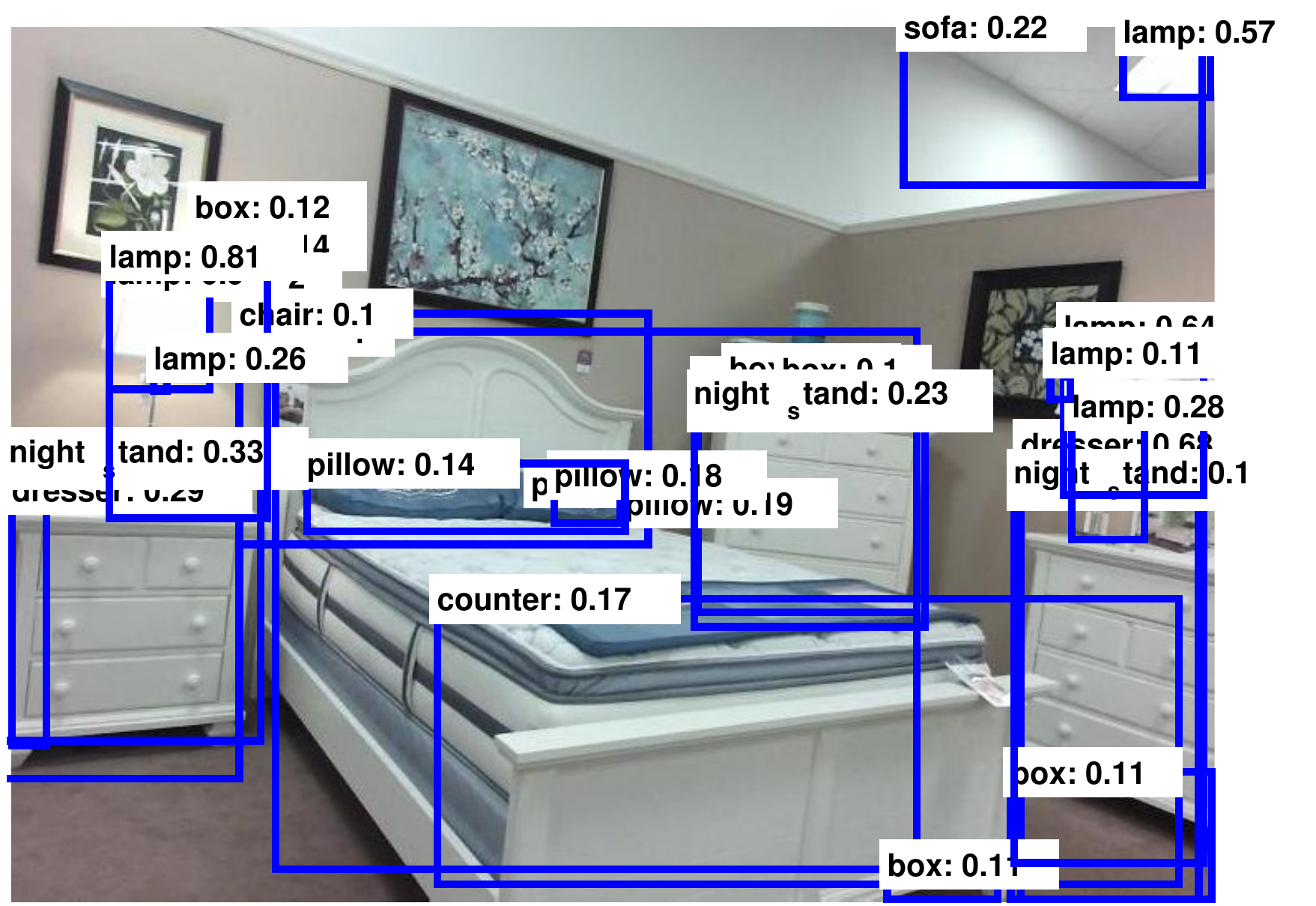}} 
  \subfigure[Context Selection]{ 
    \label{fig:subfig:b} 
    \includegraphics[width=0.30\linewidth]{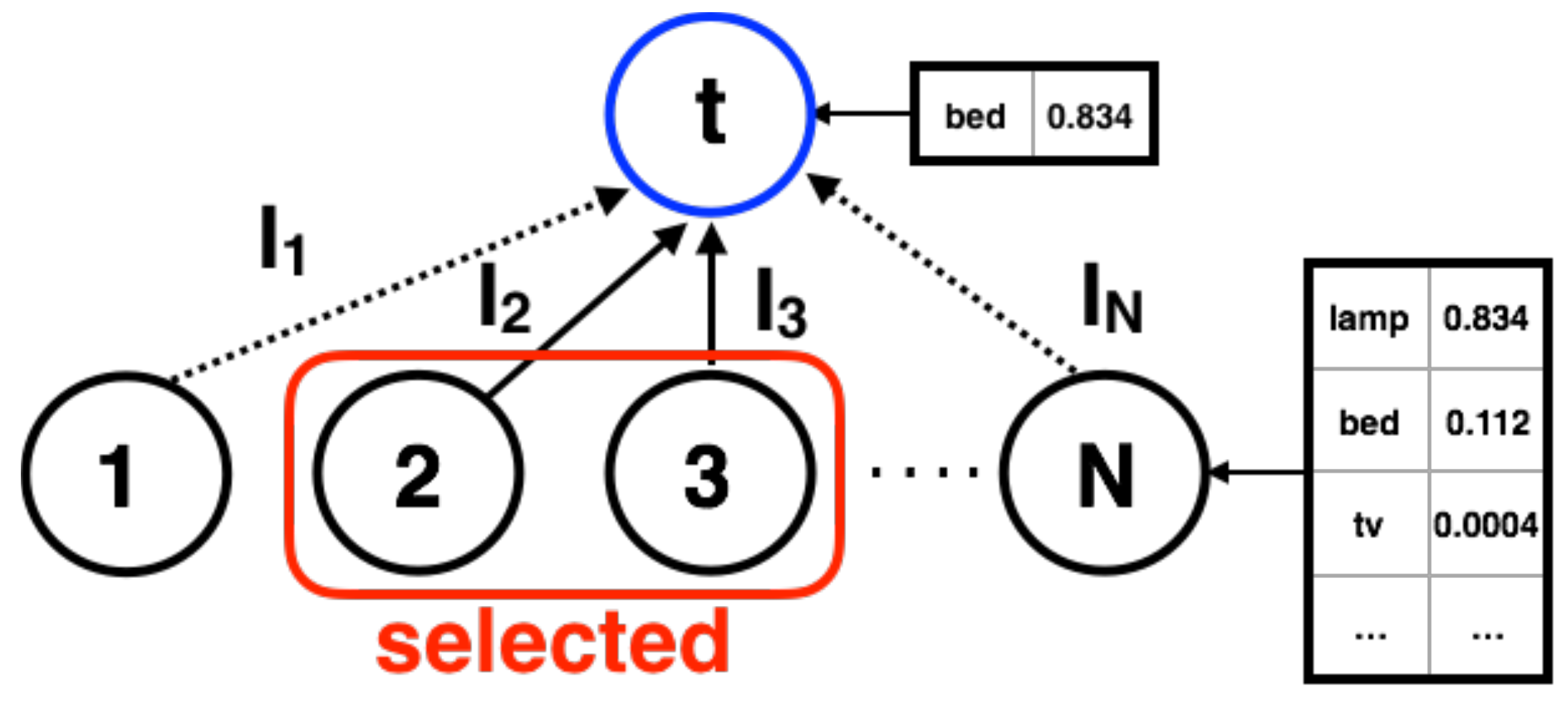}} 
    \vspace*{-5mm}
  \caption{\textbf{Context Selection with Noisy Detection:} (a) imperfect detections from the Fast R-CNN detector fine-tuned and used on the SUN RGB-D dataset produce a large number of false positives; (b) the proposed context selection method selects a subset of contextual objects from the imperfect detections to improve detection.} 
  \label{fig:subfig} 
\end{figure}

To address this question, we study possible reasons why context might not improve detection. First, imperfect extraction of context information introduces errors into contextual inference. For instance, when visual context information is extracted through imperfect appearance-based detectors, as shown in Figure \ref{fig:subfig:a}, incorrectly-detected regions can introduce noise into contextual inference, limiting the gain from context provided by correctly detected regions. Second, contextual information that is hard to extract or has low predictive power can introduce errors into context that is easy to detect and is very predictive. For example, when predicting the presence of a pillow in an image using context provided by other objects of different categories, some object categories, such as beds and sofas, which are easier to detect and have strong relationships with respect to pillow, are informative as context. Others, such as boxes and pictures, which are either hard to detect or irrelevant to pillows, are likely to be useless or even misleading. 

To further investigate these challenges, we conduct a simulation to study the role of context in isolation, without appearance-based clues. Since reliable contextual relationships between object pairs can be most reliably learned in sufficiently structured scenes, we utilize the SUN RGB-D \cite{Song_2015_CVPR} dataset, which is one of the largest indoor-scene datasets and contains a large number of annotated objects. For a given image with ground truth bounding boxes for all objects, we predict the label for each object, one at a time. The object whose label is to be predicted is referred to as the \textit{target object} and the other objects are referred to as \textit{contextual objects}. For the unknown target object, we remove the uncertainty of all remaining objects by assigning them to their ground-truth labels and use object-to-object contextual relationships to predict the label of the target object without access to its appearance. We observe very good prediction accuracy, which implies that, without detection noise, simple contextual relationships between objects can boost detection performance. We then study how predictive each object class is of a given target object class by ignoring one contextual object class at a time. The results suggest that different object classes vary in their ability to predict the presence of certain target object categories.

Motivated by these experiments, we propose a region-based context re-scoring method with dynamic context selection, illustrated in figure \ref{fig:subfig:b}, which seeks to eliminate false positive contextual regions while emphasizing likely true positive and informative ones. Specifically, we introduce a latent variable for each contextual region that determines if that region will be selected to provide context information. In practice, it is intractable to select the optimal set of contextual regions that provide the most trustworthy information when contradictory evidence exists, both \textit{for} and \textit{against} the target object having a certain class label. Instead, we decompose the problem by selecting informative regions that provide the strongest supporting and refuting evidence independently to compute a \textit{For upper-bound} and an \textit{Against upper-bound} of the confidence score, and then re-score the confidence for that object being in that class based on the difference between the two upper-bounds. The model for computing the two upper-bounds is trained by latent-SVM \cite{Felzenszwalb:2010:ODD:1850486.1850574}. The proposed method is evaluated on the SUN RGB-D dataset and achieves $48.25\%$ mean average precision (mAP), an improvement of $\sim 2.8\%$ over using object detections without context ($45.47\%$). We also conduct experiments to study the performance of the selection model. Both the simulations on pure context and the real-world experiments using the proposed selection method demonstrate the importance of object-to-object context and the gain attributed to the context selection scheme.

\section{Related Work}
Many techniques have used context to improve performance for image understanding tasks. For instance, Torralba \cite{Torralba:2003:CPO:644361.644382} proposed a framework for modeling the relationship between context and object properties, based on correlations between the statistics of low-level features across the entire scene and the objects that it contains. Divvala \textit{et al.} \cite{conf/cvpr/DivvalaHHEH09} defined several context sources and proposed a context re-scoring method that uses a regression model on multiple contextual features. Felzenszwalb \textit{et al.} \cite{Felzenszwalb:2010:ODD:1850486.1850574} proposed a simple context re-scoring model running on appearance-based detections. Graphical models have been widely applied to image segmentation and recognition tasks by jointly modeling appearance, geometry and contextual relations \cite{export:67415,DBLP:conf/cvpr/YaoFU12,Gould+al:ICCV09,conf/cvpr/ChoiLTW10}.  In \cite{Lin_2013_ICCV}, context clues were extended from 2D to 3D object detection. 

Recent advances in deep neural networks and R-CNN based detectors in both 2D and 3D have resulted in reliable appearance-based detectors \cite{DBLP:journals/corr/GirshickDDM13,DBLP:journals/corr/Girshick15,NIPS2015_5638,DBLP:journals/corr/GuptaHM15}. Context models have also been applied to deep learning features or detection results. Wenqing \textit{et al.} \cite{crf} evaluated the performance of a joint CRF model on Faster R-CNN detections. Zhang \textit{et al.} \cite{DeepContext} adapted the topology of neural networks to embed 3D context and performed holistic scene understanding based on scene context. Bell \textit{et al.} \cite{DBLP:journals/corr/BellZBG15} explored the use of spatial Recurrent Neural Networks (RNNs) to gather context information as well as to capture fine-grained details from multiple lower-level convolutional layers.

In contrast to the above approaches, our method introduces the concept of context selection to identify a subset of contextual objects which are highly likely to be true positives and informative. Our context selection method can be viewed as a "hard" attention based visual attention model that only pays attention to the selected contextual objects \cite{attention}.

\section{The Role of Pure Context}\label{pureContext}
We first conduct an experiment to analyze the utility of pure contextual relationships between objects in object detection. In this experiment, we only consider the ground-truth bounding boxes, and the label for a given box is predicted using only context information between the target box and the remaining boxes for other objects in an image. When predicting the label of the target object, we only consider its bounding box and intentionally ignore appearance information such as color, shape and texture. Moreover, the ground-truth labels and bounding boxes of all contextual objects are revealed to remove the influence of uncertainty in context. We consider three types of object-to-object context: co-occurrence, relative scale and spatial relationships.

\subsection{Predicting Object Class using Pure Contextual Relationships}
Prediction is performed by a linear classifier. Given an image $I$, assume there are $N+1$ ground-truth objects, drawn from $M$ object categories. When predicting the label of a target object $t$, the ground-truth bounding boxes of $t$ and the remaining $N$ objects are given, along with the the labels for all objects other than $t$. The confidence that object $t$ is in class $T$ is:
\begin{align}
\label{model_pure}
     & Score(o_{tT})  =  \sum^{N}_{j=1} \sum^{M}_{i=1} [  Co(o_{tT},o_{ji};\mathbf{w}_{co}) +Sc(o_{tT},o_{ji};\mathbf{w}_{sc})  +Sp(o_{tT},o_{ji};\mathbf{w}_{sp})]\cdot l_{ji}+b,
\end{align}
where $o_{tT}$ indicates that the target object $t$ is assigned label $T$,  and $o_{ji}$ indicates that the $j^{th}$ contextual object is in class $i$, $l_{ji}$ is a binary indicator variable with $l_{ji}=\textbf{1} \{ label_j = i\}$, and $b$ is a bias term. The terms $Co( \cdot )$, $Sc(\cdot)$ and $Sp(\cdot)$ measure co-occurrence, relative scale and spatial relationship defined in equation \eqref {model_CO_SC_SP}:  
\begin{align}
\label{model_CO_SC_SP}
&Co(o_{tT},o_{ji};\mathbf{w}_{co})=w_{co}(T,i) \cdot \log d_{co}(o_{tT},o_{ji}) \\ \nonumber
&Sc(o_{tT},o_{ji};\mathbf{w}_{co})=w_{sc}(T,i) \cdot \log d_{sc}(o_{tT},o_{ji}, r) \\ \nonumber
&Sp(o_{tT},o_{ji};\mathbf{w}_{sp})= w_{spx}(T,i) \cdot \log d_{spx}(o_{tT},o_{ji}, x) +w_{spy}(T,i) \cdot \log d_{spy}(o_{tT},o_{ji},y),
\end{align}
where $\mathbf{w}_{co}$, $\mathbf{w}_{sc}$, $\mathbf{w}_{sp} $ are weight vectors for each set of contextual features respectively. The term $d_{co}(o_{tT},o_{ji})$ is the likelihood that a target object $t$ of category $T$ and object $j$ of category $i$ co-occur in the same image. The term $ d_{sc}(o_{tT},o_{ji},r) $ is the likelihood that a target object $t$ of category $T$ and object $j$ of category $i$ have relative scale ratio $r$. The terms $d_{spx}(o_{tT},o_{ji}, x)$ and $d_{spy}(o_{tT},o_{ji},y)$ are the likelihoods that a target object $t$ of category $T$ and object $j$ of category $i$ have relative spatial distance (distance along one axis normalized by the height/width of the image) $x$ along X-axis, and $y$ along Y-axis, respectively. The likelihoods are learned from the statistics of the training set. For the co-occurrence context information, we use a two-bin histogram to represent the likelihood for the co-occurrence of a target-context object pair. For the relative scale and the spatial context, we categorize the relative scale ratios and relative distances into n slots and use an n-bin histogram to represent the likelihoods. Examples of the histograms can be found in the supplementary material. Laplace smoothing is applied to avoid zero counts.

Given this linear model and the features extracted based on the likelihoods, we train a multi-class classifier using structural SVM \cite{Tsochantaridis:2004:SVM:1015330.1015341}. To evaluate the performance of object detection using pure context, we compute prediction accuracies on the 19 common objects used in the SUN RGB-D dataset; the same object categories are used as contextual objects. The average accuracy is $70.68\%$, which is quite high considering that no appearance information is utilized. The accuracy for each object class can be found in the supplementary material.

\subsection{The Role of Different Contextual Object Categories}\label{importance}
The above experiment shows the predictive potential of pure context. Do all object categories provide equally informative context when predicting the label of a target object, or are some of them more informative than others? To answer this question, we evaluate the predictive power of each object category with respect to a given target object category. For each target object category $T$, we measure the relative accuracy loss (RAL), defined as $RAL(T,C,i)=\frac{Accuracy_{C-i}(T)-Accuracy_{C }(T)}{Accuracy_{C}(T)}$, when removing the $i^{th}$ object category from the set C of contextual object categories. Some RAL examples are shown in Table \ref{RAC:pillow} and \ref{RAC:book}. For each target object category, we show the object categories that lead to the top five largest RALs. We observed that different object categories have significantly different predictive power depending on certain object categories.
\begin{table}[htp]\scriptsize
\floatsetup{floatrowsep=qquad,captionskip=5pt} \tabcolsep=9pt
\begin{floatrow}
\ttabbox{\caption{Relative Accuracy Loss (RAL): $T$ = Pillow}\label{RAC:pillow}}{%
\begin{tabular}{p{0.4cm}<{\centering} p{0.4cm}<{\centering} p{0.4cm}<{\centering} p{0.4cm}<{\centering} p{0.4cm}<{\centering} p{0.4cm}<{\centering}}\hline
   &  pillow& 	bed	& sofa &	lamp	& night-stand  \\
   \hline
   RAL & 0.28	&0.24&	0.17	&0.08&	0.04   \\
 \hline
\end{tabular}}
\ttabbox{\caption{Relative Accuracy Loss (RAL): $T$ = Bookshelf}\label{RAC:book}}{%
\begin{tabular}{p{0.4cm}<{\centering} p{0.4cm}<{\centering} p{0.4cm}<{\centering} p{0.4cm}<{\centering} p{0.4cm}<{\centering} p{0.4cm}<{\centering}}\hline
   &  book-shelf& chair& desk&table& box  \\
   \hline
   RAL & 0.35	&0.27 &	0.17	&0.11&	0.03   \\
 \hline
\end{tabular}}
\end{floatrow}
\end{table}

In summary, pure contextual information between object pairs has high predictive power, but each contextual-object category, not surprisingly, predicts some target categories much better than others.

\section{Region-based Context Re-scoring with Dynamic Context Selection}
Based upon the above analysis, we propose a model to improve detection based on context, where contextual objects are detected automatically and are thus noisy probabilistic detections. We utilize the appearance clues from state-of-the-art detectors for predicting a target object's label. The same contextual relationships discussed in the previous analysis between a detected target bounding box and the remaining ones are utilized, but each box is a region with an $(M+1)$ associated probability distribution over the possible labels including the background. We introduce binary latent variables for all contextual regions, indicating whether a contextual region is selected in the context re-scoring process. 

\subsection{Test-Time Re-scoring using For-and-Against Upper Bounds}
We propose a region-based context re-scoring model with context selection. The re-scored confidence for the target object $t$ being in category $T$ is:
\begin{align}
\label{model_select}
     Score(o_{tT})  = & w_0\log A(o_{tT})+\sum^{N}_{j=1} \sum^{M}_{i=1} [  Co(o_{tT},o_{ji};\mathbf{w}_{co})  +Sc(o_{tT},o_{ji};\mathbf{w}_{sc}) +Sp(o_{tT},o_{ji};\mathbf{w}_{sp}) \\ \nonumber     
     &+w_{A_c}(i)\log A_c(o_{ji})  ] \cdot l_j+b,
\end{align}
where $A(o_{tT} )$ and $A_c(o_{ji})$ represent the appearance-based confidence scores of the target and the contextual objects, $w_0$ and $w_{A_c}(i)$ are the corresponding weights, and $l_j$ is the binary indicator variable for context selection. The proposed method can be viewed as a tree model where the target object (the root) collects context information from the selected contextual objects (the leaves). In contrast to traditional graphical models, the proposed method is an approximation that decomposes the re-scoring process into two independent ones due to the intractability of jointly solving the context selection with contradictory context information. 

Our model, intuitively, corresponds to a courtroom where the prosecutors tries to prove that the target object does not come from a given class by providing the most compelling negative evidence (a small set of confidently detected context objects whose presence is inconsistent with that label for the target object), while the defendant's lawyer provides the most compelling evidence for the truth of the claim that the target object is from the given class.  Our goal is to learn, from training data, how these "arguments" should be constructed for a given target class.  That is, we seek to learn a computational model for a multi-valued logic \cite{Ginsberg88multivaluedlogics:} in an attempt to avoid noise in detection and ambivalence in contextual prediction (from the large number of incorrect and irrelevant potential objects in an image) from overwhelming the clear and compelling evidence concerning the identity of the target object.  This type of multi-valued logic has been used before for human detection (where reasoning considered occlusion and image border effects \cite{Bilattice}), but actually learning how to choose these evidential arguments from training data has not been done before. So, our solution involves identifying the different sources that provide supporting and refuting evidence independently, and then to combine the \textit{degree of belief for} and the \textit{degree of belief against} to obtain the final confidence of a target object being in class $T$. Specifically, we first re-score each target object $t$ by selecting the evidence that most strongly supports it being in a certain class to compute its \emph{For-Score}, and select the evidence that most strongly argues against it for its \emph{Against-Score}. Both the \emph{For-Score} and the \emph{Against-Score} can be computed by maximizing function \eqref{model_select} over all possible indicator vectors that consist of indicator variables for all contextual regions, but with different weight vectors. The weight vector for computing the \emph{For-Score} is learned with positive samples that are in class $T$, while the weight vector for the \emph{Against-Score} considers the objects that are not in class $T$ as the positive samples. The \emph{For-Score} can be viewed as a belief upper bound for a target object being in class $T$. In both cases we select contextual regions with high appearance-based confidence scores by forcing the weight $w_{A_c}(i)$ to be positive. The final degree of belief for the target object $t$ being in class $T$ is the margin between the \textit{For-and-Against upper bounds}:  $Score(o_{tT})=Score_{For}(o_{tT})-Score_{Against}(o_{tT})$.

\vspace*{-1mm}
\subsection{Training with Latent-SVM}
The proposed re-scoring model with dynamic context selection can be trained using latent-SVM \cite{Felzenszwalb:2010:ODD:1850486.1850574}. The processes for learning the weights for computing the \emph{For-Score} and the \emph{Against-Score} are the same except for the choice of positive samples. We describe the general training process. The weight vector and feature vector for an sample x are shown in equation \ref{weight} and \ref {feature}.

\begin{equation}
\label{weight}
\mathbf{w}=\left \{ w_0, \mathbf{w}_{co},\mathbf{w}_{sc},\mathbf{w}_{sp},\mathbf{w}_{A_c}, b\right \},
\end{equation}

\begin{align}
\label{feature}
     &\mathbf{\phi}(x,\mathbf{l})= \{ \log A(o_{tT}),  \sum^{N}_{j=1}  \log d_{co}(o_{tT},o_{j1})\cdot l_j,  \\ \nonumber
    &    \cdots,  \sum^{N}_{j=1} \log d_{co}(o_{tT},o_{jM})\cdot l_j ,  \sum^{N}_{j=1} \log d_{sc}(o_{tT},o_{j1}) \cdot l_j,    \\ \nonumber 
    &  \cdots,  \sum^{N}_{j=1}  \log d_{sc}(o_{tT},o_{jM})\cdot l_j, \sum^{N}_{j=1}  \log d_{sp}(o_{tT},o_{j1})\cdot l_j, \\ \nonumber 
    &   \cdots, \sum^{N}_{j=1}   \log d_{sp}(o_{tT},o_{jM})\cdot l_j, \sum^{N}_{j=1} \log A_c(o_{j1}) \cdot l_j \cdots, \sum^{N}_{j=1} \log A_c(o_{jM}) \cdot l_j, 1 \},
\end{align}

where $\mathbf{l}$ is the vector consists of indicator variables.

The re-scored confidence for a sample x is determined by a classifier using a function of form:

\begin{equation}
\label{classifier}
f_{\mathbf{w}}(x)=\max_{\mathbf{l} \in \mathbf{L} (x) } \mathbf{w}  \cdot   \mathbf{\phi}(x,\mathbf{l}),
\end{equation}
where $\mathbf{L} (x)$ consists of all possible latent vectors for sample x. The objective function to be minimized is:

\begin{equation}
\label{loss}
loss(\mathbf{w})= \frac{1}{n}  \sum^{n}_{i=1} \max (0,1-y_i f_{\mathbf{w}}(x_i)) +\frac{\lambda}{2}\|\mathbf{w}\|^2,
\end{equation}
where we adopt the hinge loss to minimize the loss in a max-margin manner. The constant $ \lambda $ is used to weight the regularization term.

Although a latent-SVM leads to a non-convex optimization, we can efficiently solve it using coordinate descent by leveraging its semi-convexity property. The coordinate descent method involves two steps. Firstly, positive samples are relabeled by selecting contextual-regions that scores the target object highest by solving a linear programming problem:
\begin{equation}
\label{relabeling}
\mathbf{l}=\text{argmax}_{\mathbf{l} \in \mathbf{L} (x)} \mathbf{w}  \cdot   \mathbf{\phi} (x,\mathbf{l}),
\end{equation}
and then, the weight vector $\mathbf{w}$ is optimized by solving a convex optimization problem by minimizing the loss in equation \eqref{loss} given the relabeled positive samples.

\section{Experiments}
\subsection{Dataset}
We use the SUN RGB-D dataset, which contains images from \cite{Silberman:ECCV12,Janoch2011,SUN3D}, to evaluate our proposed method. We consider the 19 most common classes in the dataset. The performance is evaluated through the average precision (AP) of object detection. For comparison, we evaluated several R-CNN based detectors, and chose as the baseline one that utilizes the depth modality by supervision transfer (ST) \cite{DBLP:journals/corr/GuptaHM15}, which uses object proposals from \cite{DBLP:journals/corr/GuptaGAM14} and yields state-of-the-art mAP on the SUN RGB-D dataset for 2D object detection. 
\subsection{Context Selection Model and Baseline Models}
Besides ST, we also compare our context selection model (CS) with the baselines including the one that "selects all" (SA) contextual regions and the one that only considers either the \textit{For upper-bound} (FUB) or the \textit{Against upper-bound} (AUB). For each object category T, we train the model based upon appearance-based detections using latent-SVM. When predicting a target object's label, we set a precision threshold (and choose corresponding appearance-based confidence score thresholds of all contextual objects), and only consider detections with scores higher than the thresholds as potential contextual objects to ensure the context precision for the potential contextual objects of each class reaches the precision threshold. To train the FUB model, we label boxes that have ground-truth labels in class $T$ as positive samples and select from supporting evidence to obtain the \textit{For upper-bound}. The training process for the AUB model is obtained by simply reversing the positive and negative training labels. During the test, for each target object $t$, the context selection model selects supporting refuting evidence separately to compute the FUB and the AUB, and then uses the margin between them as the final confidence score for object $t$ being in class $T$. 

\textbf{Does context selection work?} We compare the context selection model with the select-all model by measuring average precision. Table \ref{tab:AP} shows the average precision of the 19 object classes when the precision threshold for contextual regions is set as 0.4. The Precision-Recall (PR) curves can be found in the supplementary material. We achieve only a 0.43\% mAP gain by the SA model, with some classes improving notably (counter, desk, lamp and pillow), while others (bathtub, chair, monitor, dresser, sofa, sink and toilet) degrading when considering all contextual regions. When we apply context selection, we see large improvement in AP on almost all classes compared to both the ST and SA baselines. We observe $\sim2.8\%$ mAP gain by the context selection method.

\begin{table*}[h!]\tiny
  \centering
  \vspace*{-1.5mm}
  \caption{\textbf{Average Precision (AP) on SUN RGB-D Test Set.} ST \cite{DBLP:journals/corr/GuptaHM15} is the appearance-based detector using supervision transfer. SA is the context model that selects all detections from ST as contextual regions. FUB is the context selection model that only considers supporting evidence. AUB is the one similar FUB but only considers refuting evidence. CS is the full model that selects a subset of detections as contextual regions and models the supporting and refuting evidence together. The performance gain by selecting all contextual regions is marginal, but when context selection is introduced, the performance is significantly boosted.}
  \label{tab:AP} 
  \begin{tabular}{p{0.55cm}<{\centering} p{0.18cm}<{\centering} p{0.18cm}<{\centering} p{0.18cm}<{\centering} p{0.18cm}<{\centering}  p{0.18cm}<{\centering} p{0.18cm}<{\centering} p{0.18cm}<{\centering} p{0.18cm}<{\centering} p{0.18cm}<{\centering} p{0.18cm}<{\centering} p{0.18cm}<{\centering} p{0.18cm}<{\centering} p{0.18cm}<{\centering} p{0.18cm}<{\centering} p{0.18cm}<{\centering} p{0.18cm}<{\centering} p{0.18cm}<{\centering} p{0.18cm}<{\centering} p{0.18cm}<{\centering} p{0.18cm}<{\centering} }
   \hline
   & bathtub & bed & book-shelf & box & chair & counter & desk & door & dresser & garbage bin & lamp & monitor & night stand & pillow & sink & sofa & table & tv & toilet & mAP\\
 \hline
  ST \cite{DBLP:journals/corr/GuptaHM15} &  67.70 & 76.44 & 43.45 & 18.02 & \textbf{42.15 }& 32.06 & 24.94 & 22.93 & 40.27 & 52.83& 49.73 & 47.80 & 56.30 & 48.75 & 19.92 & 50.82 & 42.83 & 45.66 & 81.35 & 45.47 \\
   \hline
   SA  &  59.61 & 77.62 & 42.12 & 19.46 & 40.01 & \textbf{35.80} & \textbf{33.86 }& 23.75 & 39.72 & 50.81 & 51.19 & 43.87 & \textbf{61.44} & 51.20 & 17.61 & 49.37 & 45.86 & 48.39 & 80.37 & 45.90 \\
   \hline
   FUB& 67.75 & 78.89 & 45.60 & 22.11 & 39.19 & 34.82 & 29.54 & 23.92 & \textbf{41.19} & 52.51 & 53.22 & 48.76 & 59.61 & 53.06 & 23.19 & 50.72 & 43.74 & 50.68 & 81.53 & 47.37 \\
   \hline
   AUB & 67.54 & 78.77 & 45.64 & 20.42 & 40.21 & 35.06 & 26.80 & 23.75 & 41.04 & 52.97 & 52.84 & 48.93 & 58.62 & 51.69 & 23.32 & \textbf{52.07} & 46.65 & 45.59 & 81.51 & 47.02 \\
   \hline
   CS   & \textbf{69.15} & \textbf{80.09} & \textbf{45.94} & \textbf{22.33} & 42.04 & 35.57 & 29.84 & \textbf{24.44} & 40.85 & \textbf{53.51} & \textbf{53.67} & \textbf{48.96} & 60.18 & \textbf{54.19} & \textbf{24.60} & 48.99 & \textbf{48.86} & \textbf{51.06}& \textbf{82.50} & \textbf{48.25} \\
   \hline  
  \end{tabular}
\end{table*}

\begin{figure}[t!]\tiny
\begin{center}
\includegraphics[width=2in]{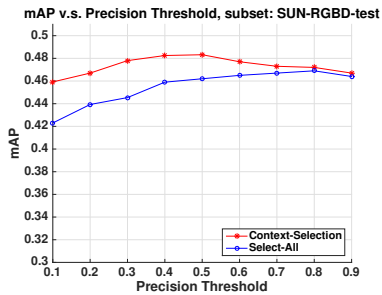}
\caption{\textbf{mAP v.s. Precision Threshold.} Our proposed approach of contextual object selection outperforms the straightforward approach of selecting the most confident contextual objects by thresholding their confidence scores at various precision targets.}
\label{fig:mAP_thres}
\end{center}
\vspace{-1.5em}
\end{figure}

\textbf{How does context selection compare to simple threshold-based filtering?} One may wonder whether the selection method can be modeled by a simple threshold-based selection on contextual regions based on their appearance-based confidence scores. We conducted an experiment to compare the performance of the proposed context selection method with that of the select-all method augmented with the choice of different precision thresholds for contextual regions. The results are shown in Figure \ref{fig:mAP_thres}. We observe that with the increase of precision threshold for contextual regions, the select-all method does perform better due to reduced noise from contextual regions. The context selection method also consistently improves as the precision threshold raises is raised from 0.1 to 0.5. Performance drops when the precision threshold exceeds 0.5. At this point, too few relevant regions survive the precision threshold. Generally, the context selection method outperforms the select-all method with precision-threshold-based filtering.

\textbf{Does the margin between \textit{For-and-Against upper-bounds} help?} Performance of context selection methods that use only one of the \textit{For upper-bound}, the \textit{Against upper-bound} and the margin between them are shown in Table \ref{tab:AP}. By ignoring the against evidence in the FUB method, the confidence scores of true positives increase as expected, but the false positives are also boosted higher. As we subtract the AUB from the FUB, we introduce refuting evidence to balance the boosting effect, and reduce false positives. We observe a performance gain of about $\sim1.0\%$ by combining the two upper-bounds together. 


\textbf{Does the selection model do more than select true positive contextual regions?} Section \ref{importance} illustrated the differential predictive power of contextual objects for a certain target object. Ideally the context selection method should select contextual objects that exist in the image and also have strong predictive power. To test that this is indeed occurring, we compare to the setting where an oracle labels the true positive contextual objects for the model to choose from. We show the APs for the SA and the CS methods with oracle in both training and testing phases in Table \ref{tab:oracle}  and label them as SA-O and CS-O, respectively. For each target object class and a given contextual object class $i$, we show the ratio between the counts of selected true positive contextual objects in class $i$ and the total number of contextual objects in that class as the selecting ratio. The top five contextual objects that have the highest selecting ratios of two target object classes are shown in Table \ref{select:pillow} and \ref{select:book}. We notice that the selecting ratios vary for different contextual object categories, and by selecting a subset of informative contextual objects, the CS-O method outperforms the SA-O method. The performance of CS-O is the upper-bound of the context selection model, and the proposed selection model is a good approximation to the upper-bound.

\begin{figure}[!h]
    \centering
    {\includegraphics[width=0.2\linewidth]{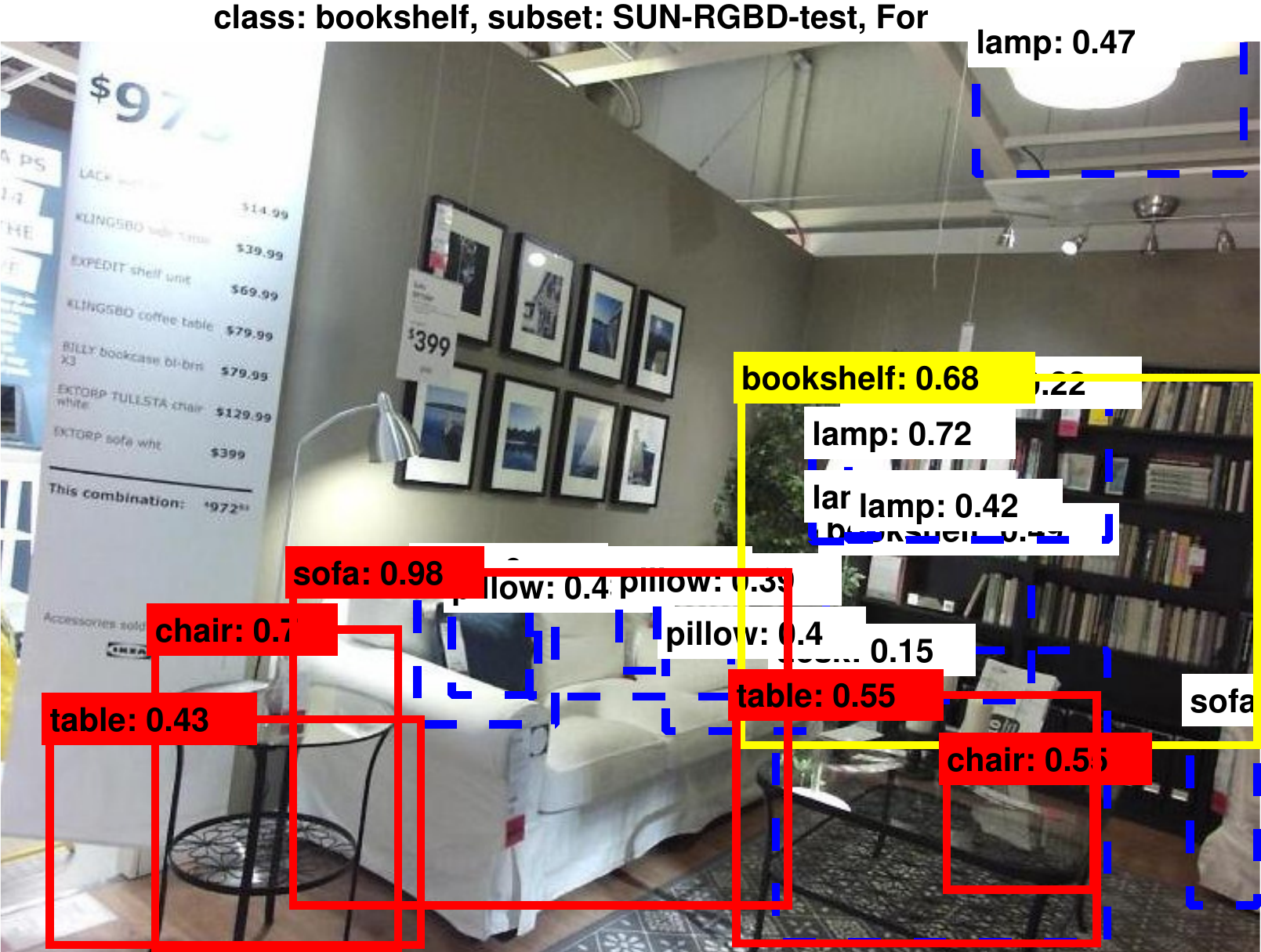}\label{fig:132_bookshelf_19}}
    {\includegraphics[width=0.2\linewidth]{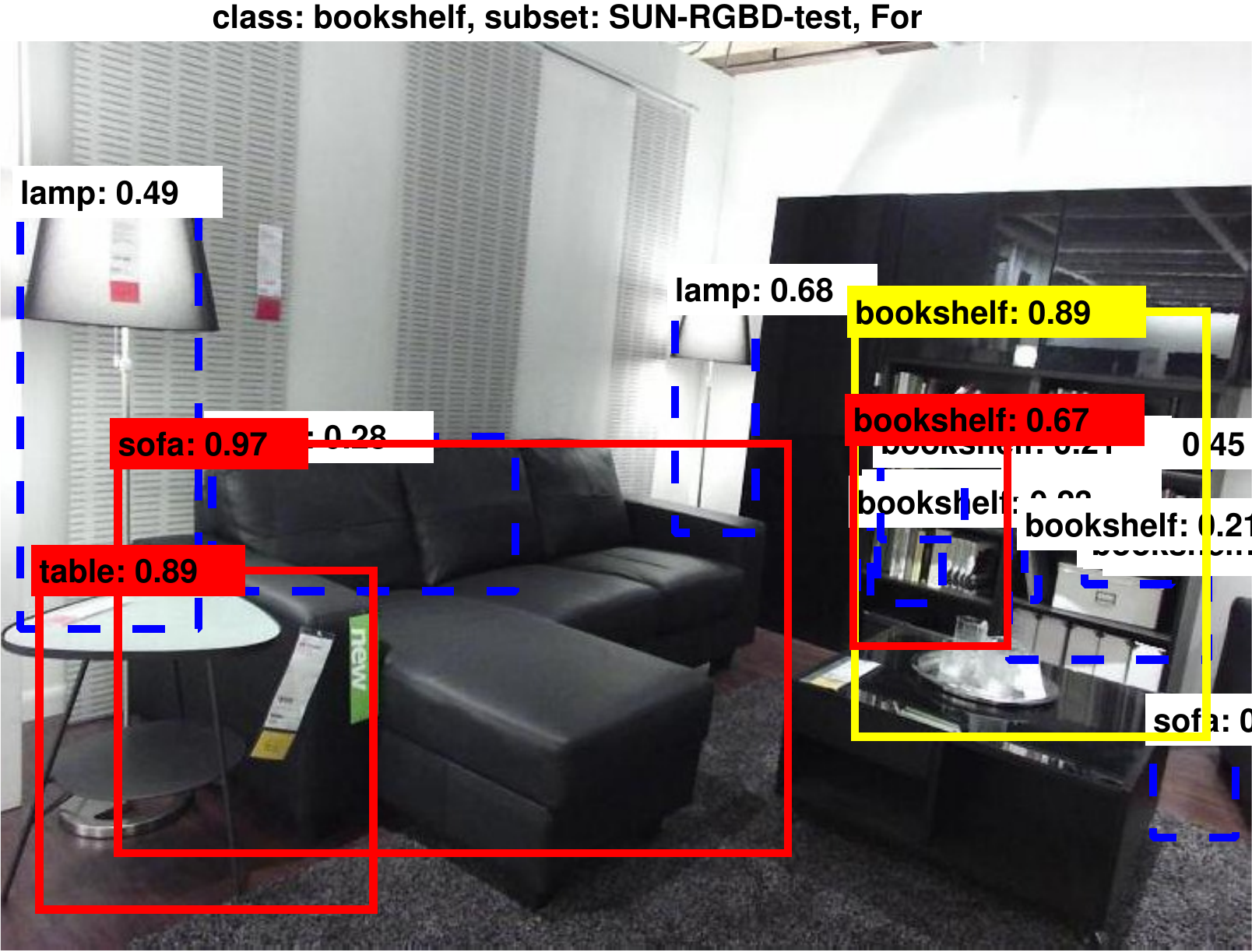}\label{fig:141_bookshelf_10}}
    {\includegraphics[width=0.2\linewidth]{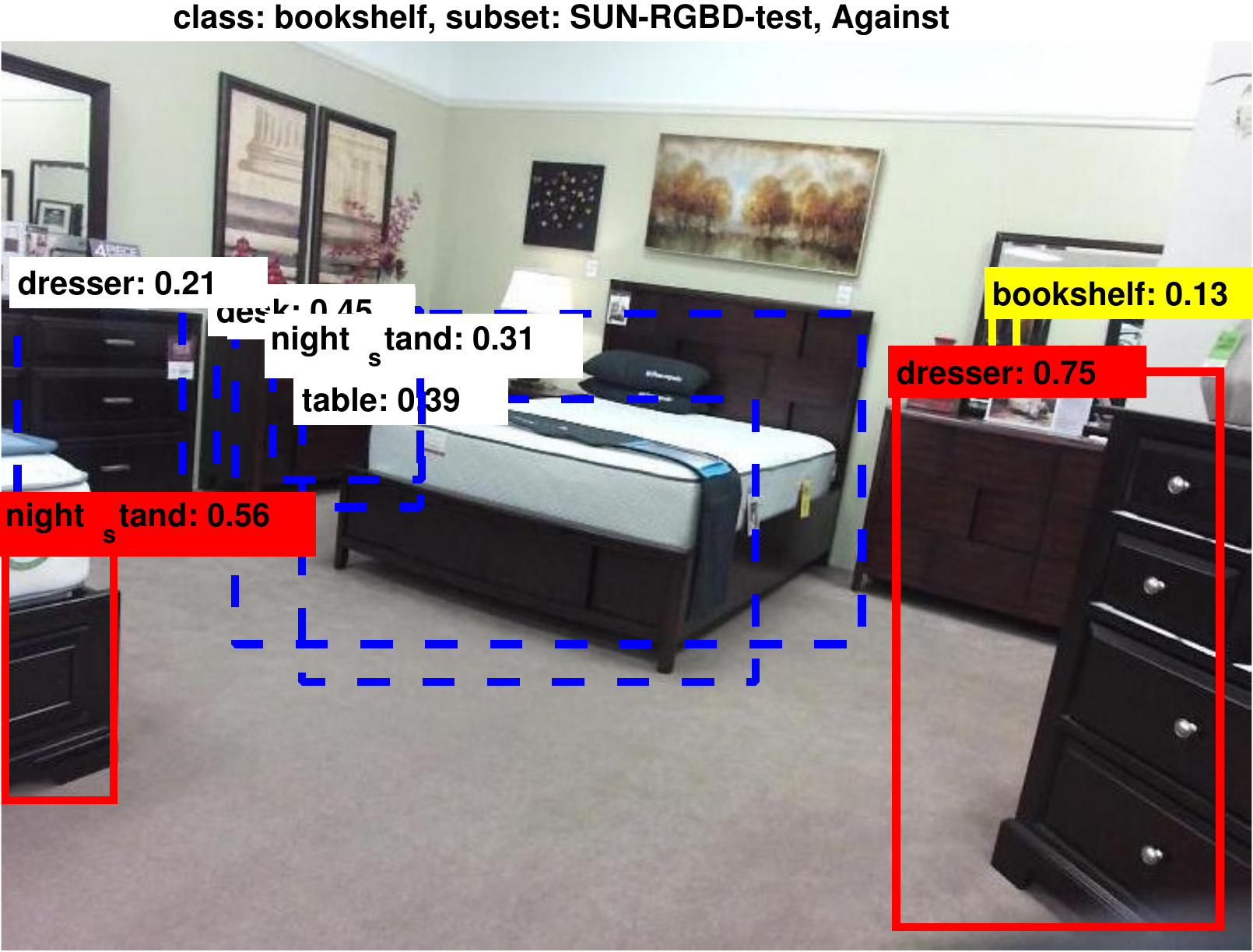}\label{fig:31_bookshelf_5}}
    {\includegraphics[width=0.2\linewidth]{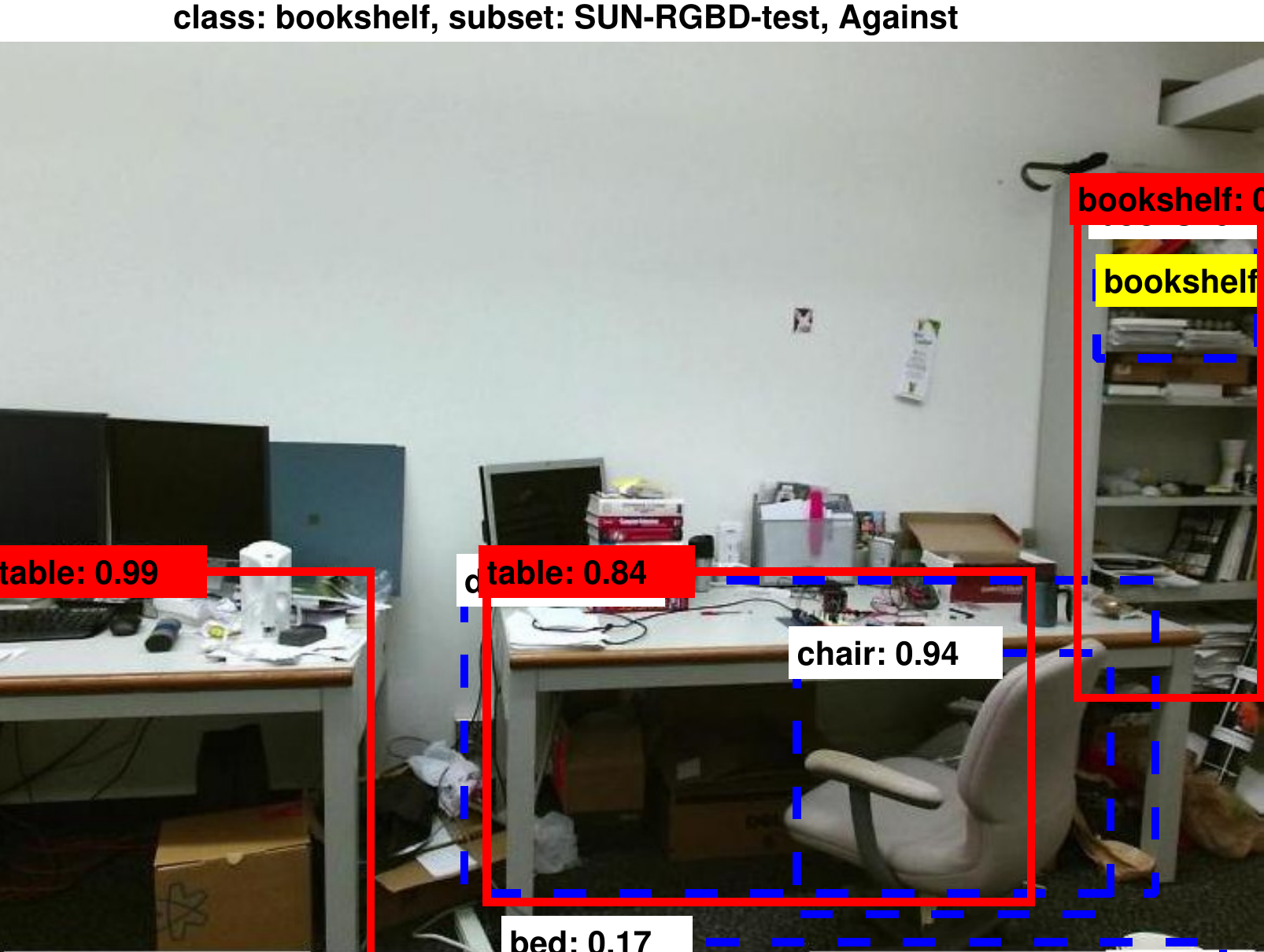}\label{fig:71_bookshelf_24}} 
    
    {\includegraphics[width=0.2\linewidth]{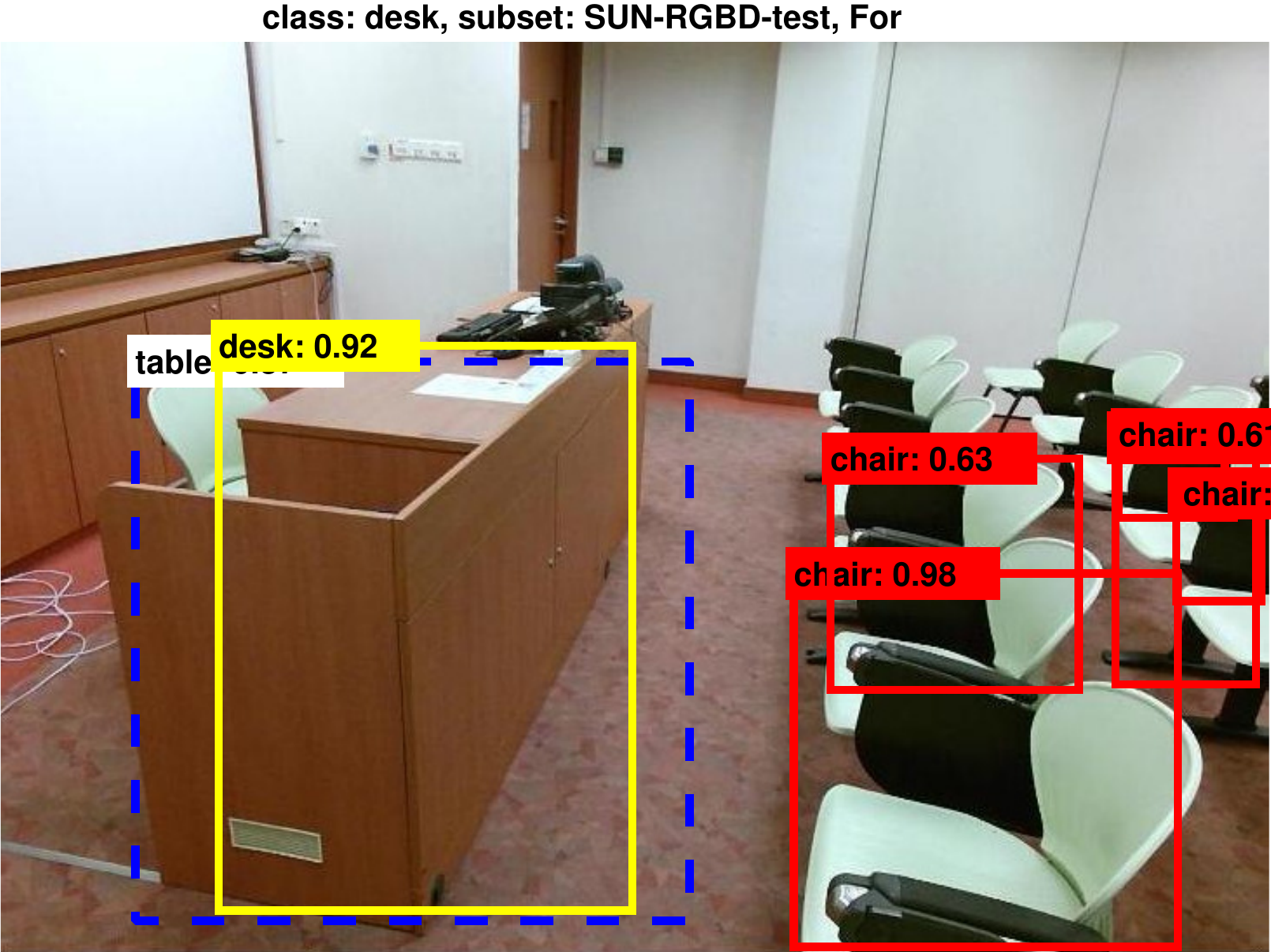}\label{fig:1205_desk_60}}
    {\includegraphics[width=0.2\linewidth]{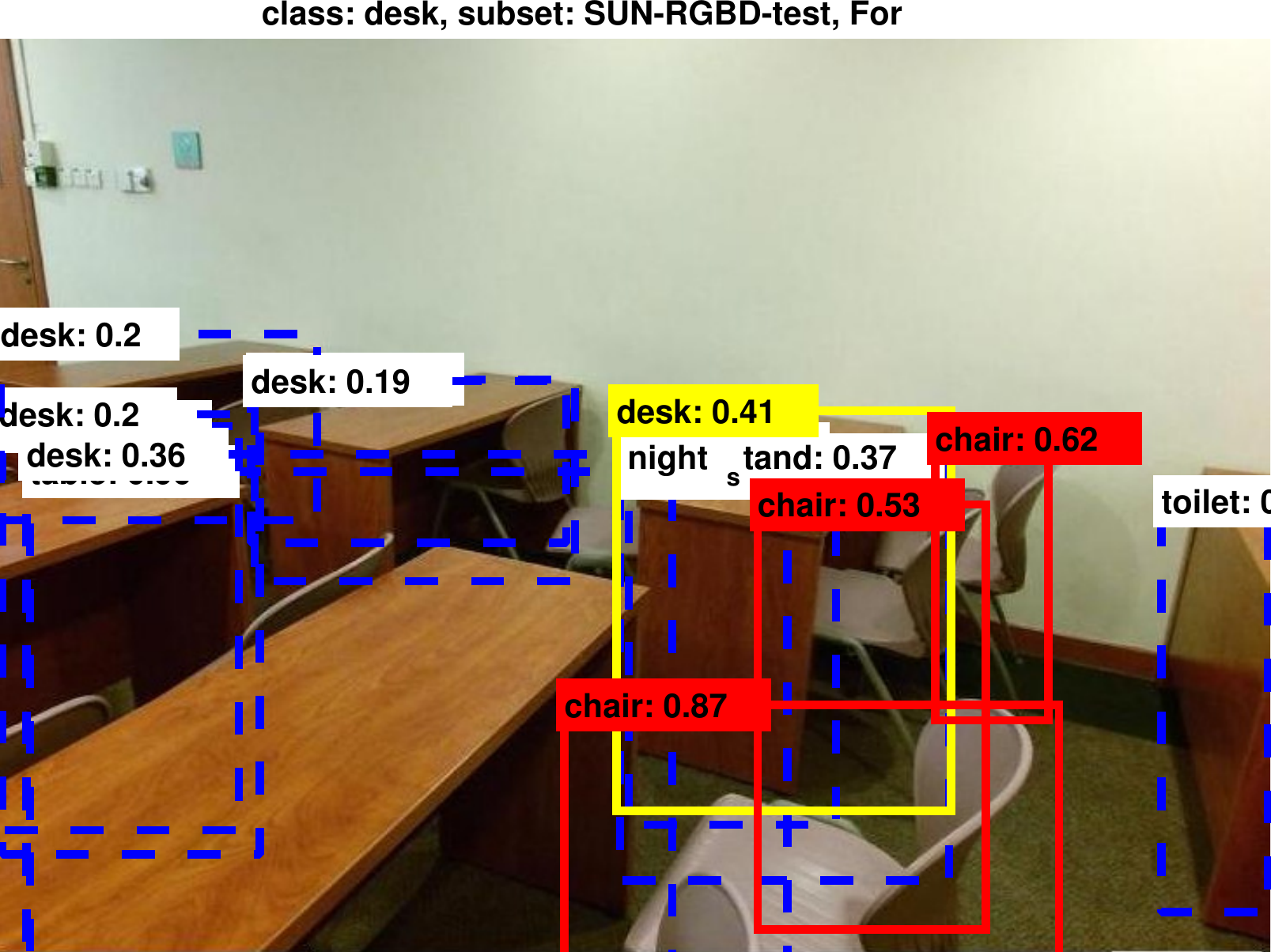}\label{fig:1249_desk_26}}
    {\includegraphics[width=0.2\linewidth]{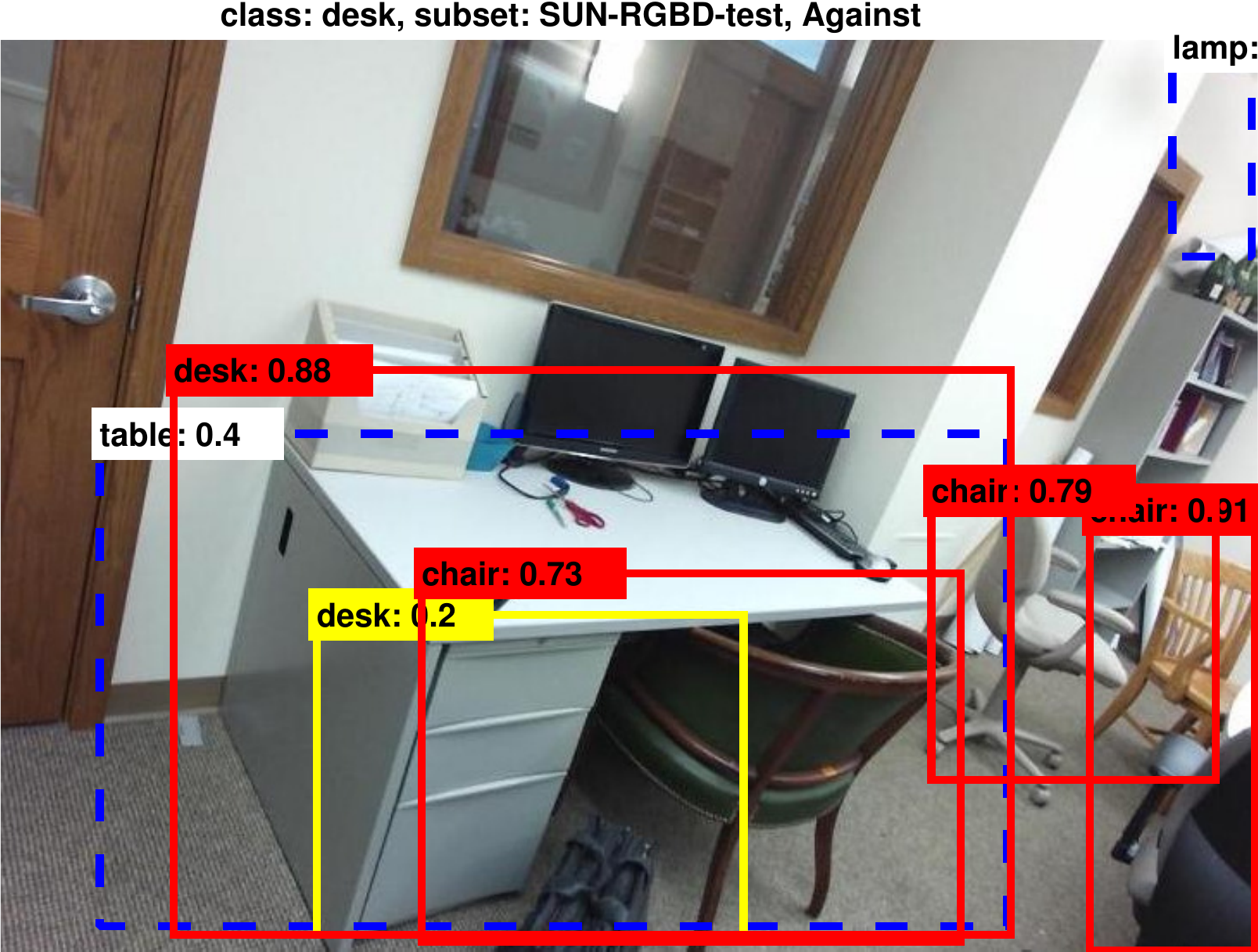}\label{fig:81_desk_22}}
    {\includegraphics[width=0.2\linewidth]{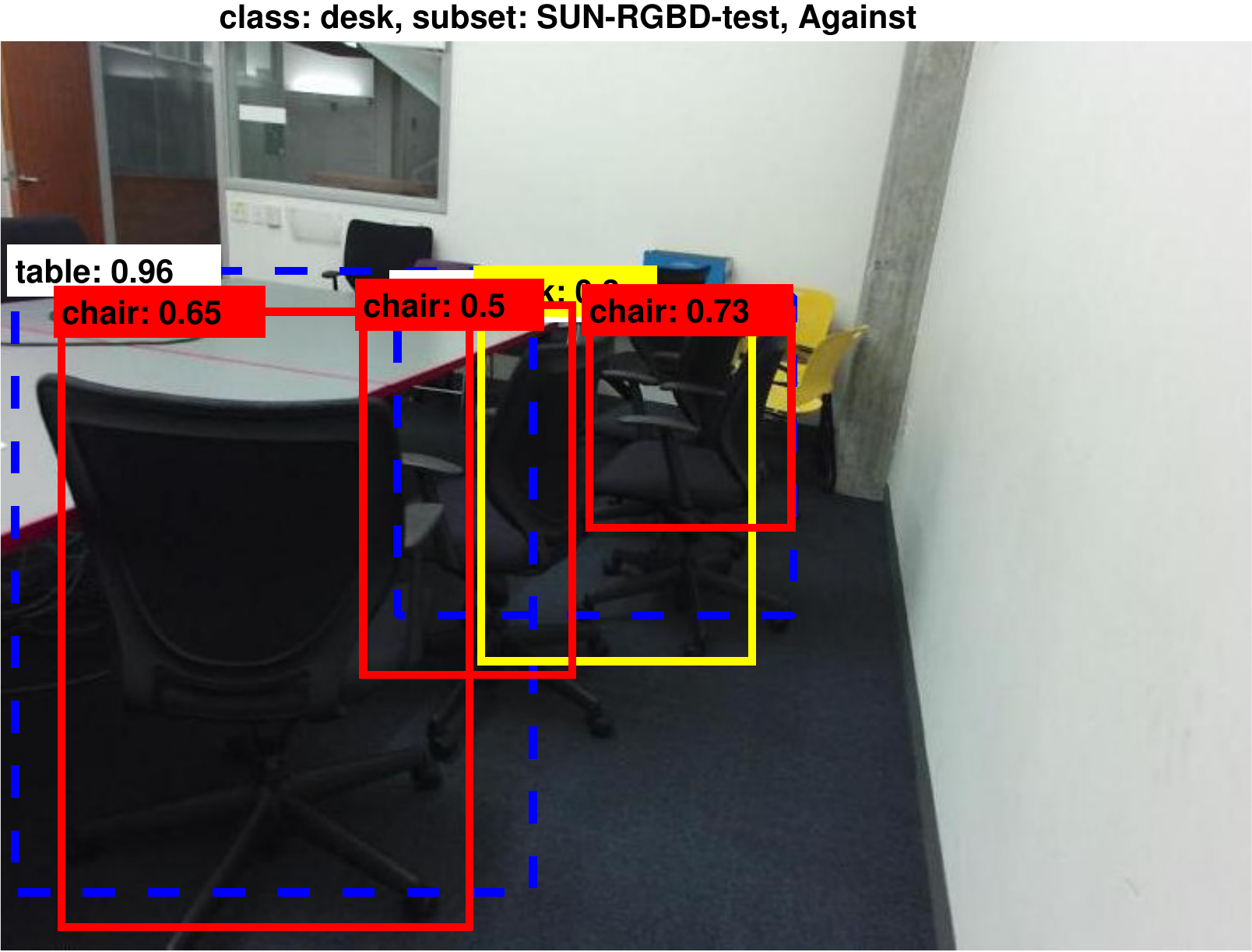}\label{fig:186_desk_31}} 
    
    {\includegraphics[width=0.2\linewidth]{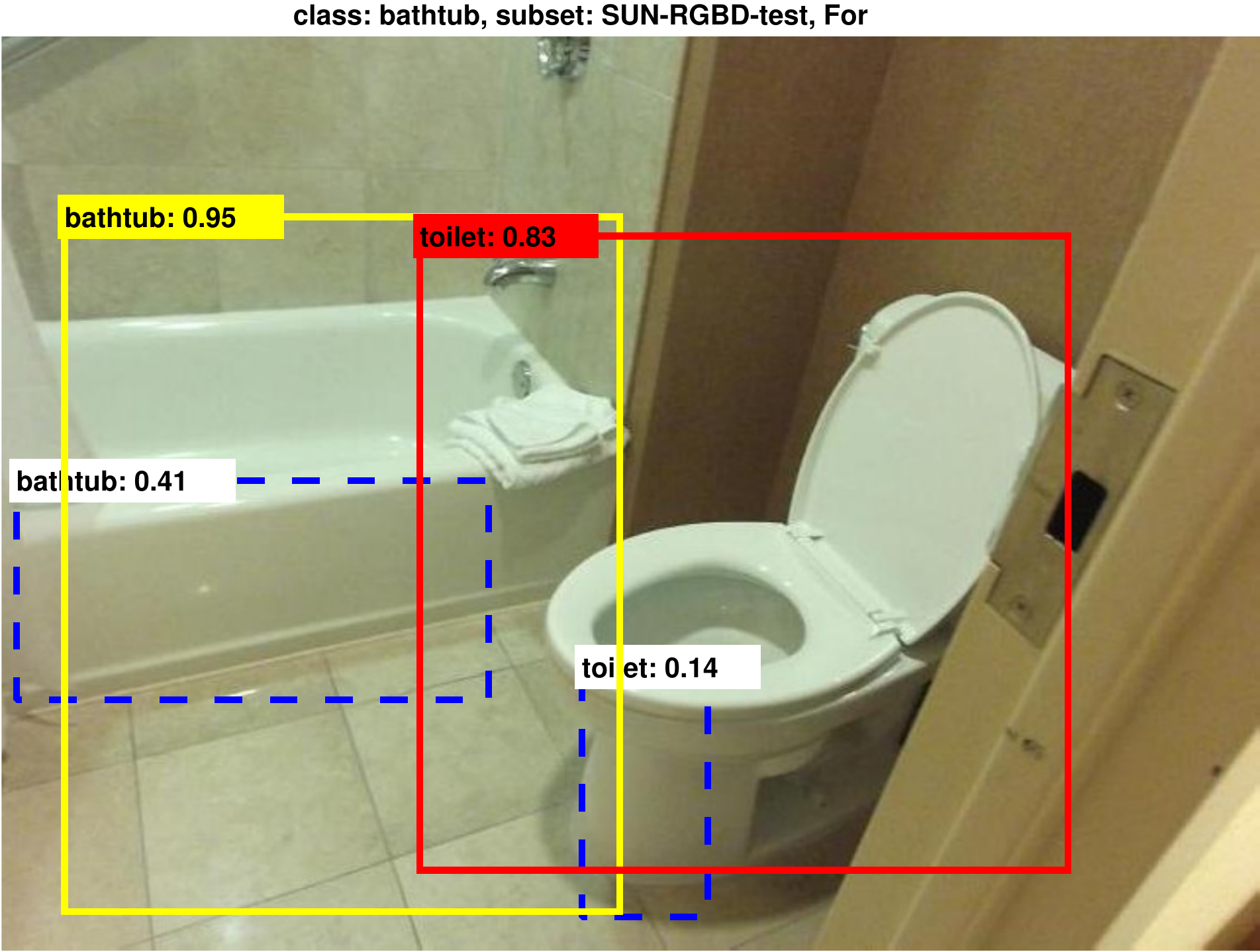}\label{fig:414_bathtub_94}}
    {\includegraphics[width=0.2\linewidth]{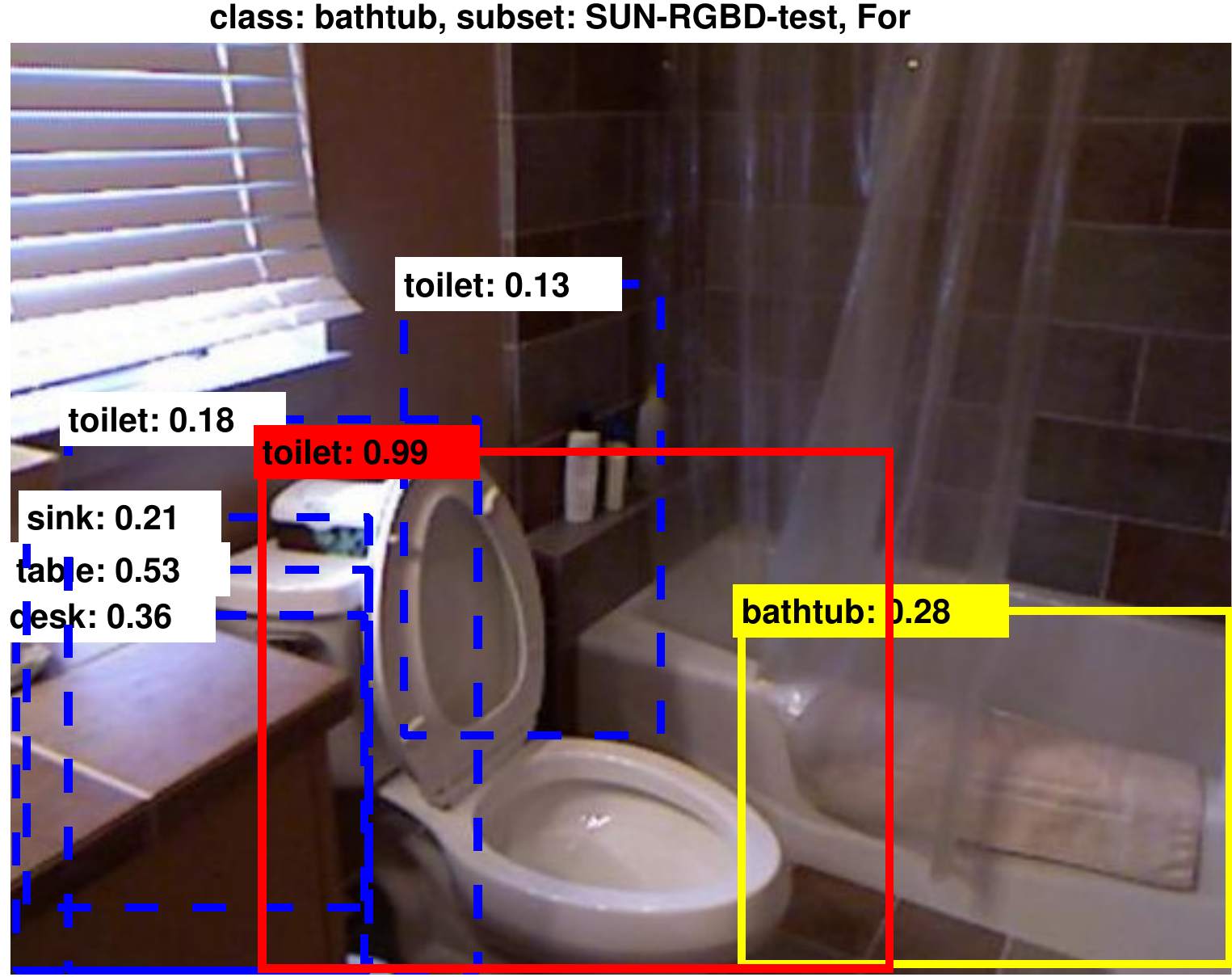}\label{fig:2176_bathtub_60}}
    {\includegraphics[width=0.2\linewidth]{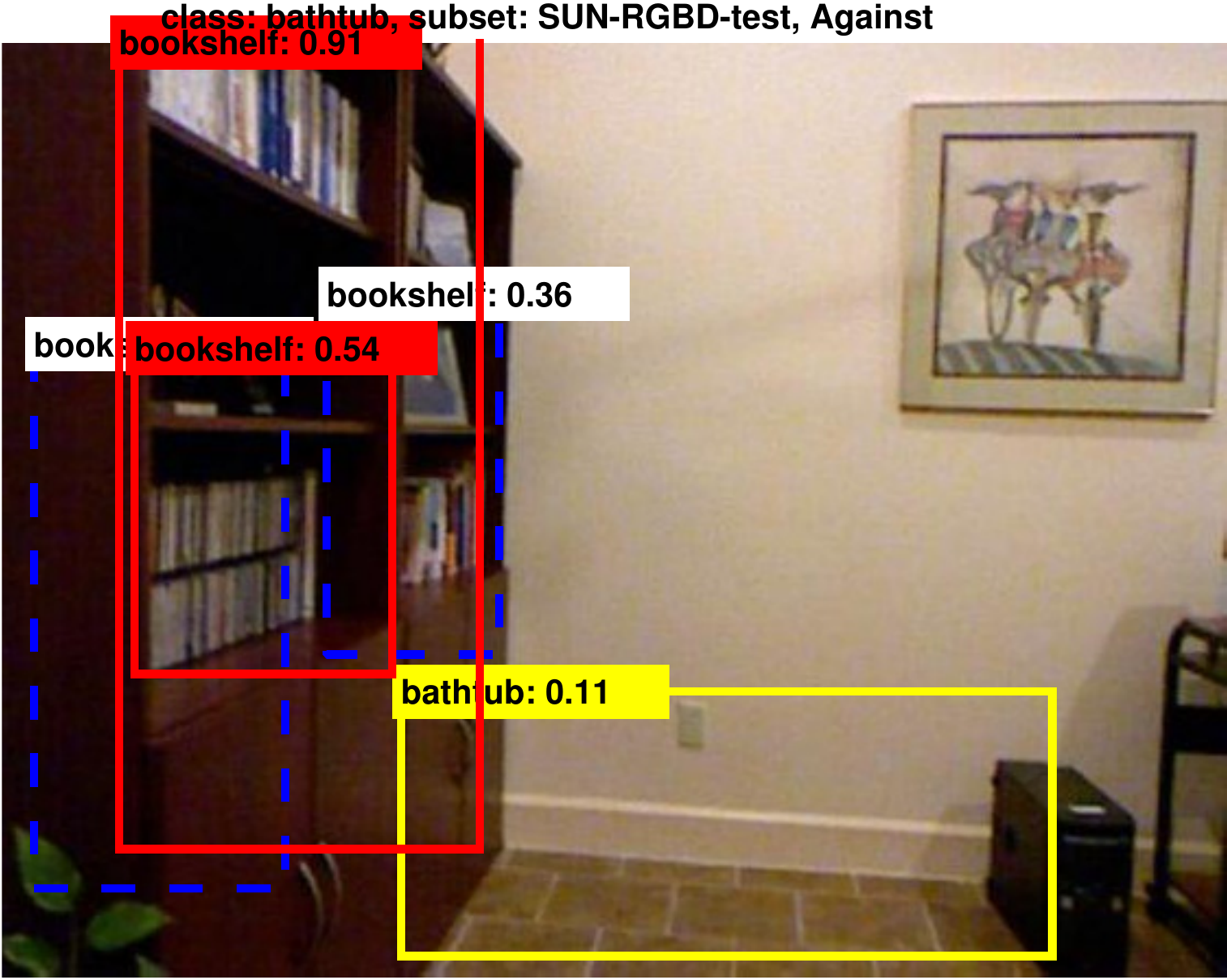}\label{fig:2048_bathtub_103}}
    {\includegraphics[width=0.2\linewidth]{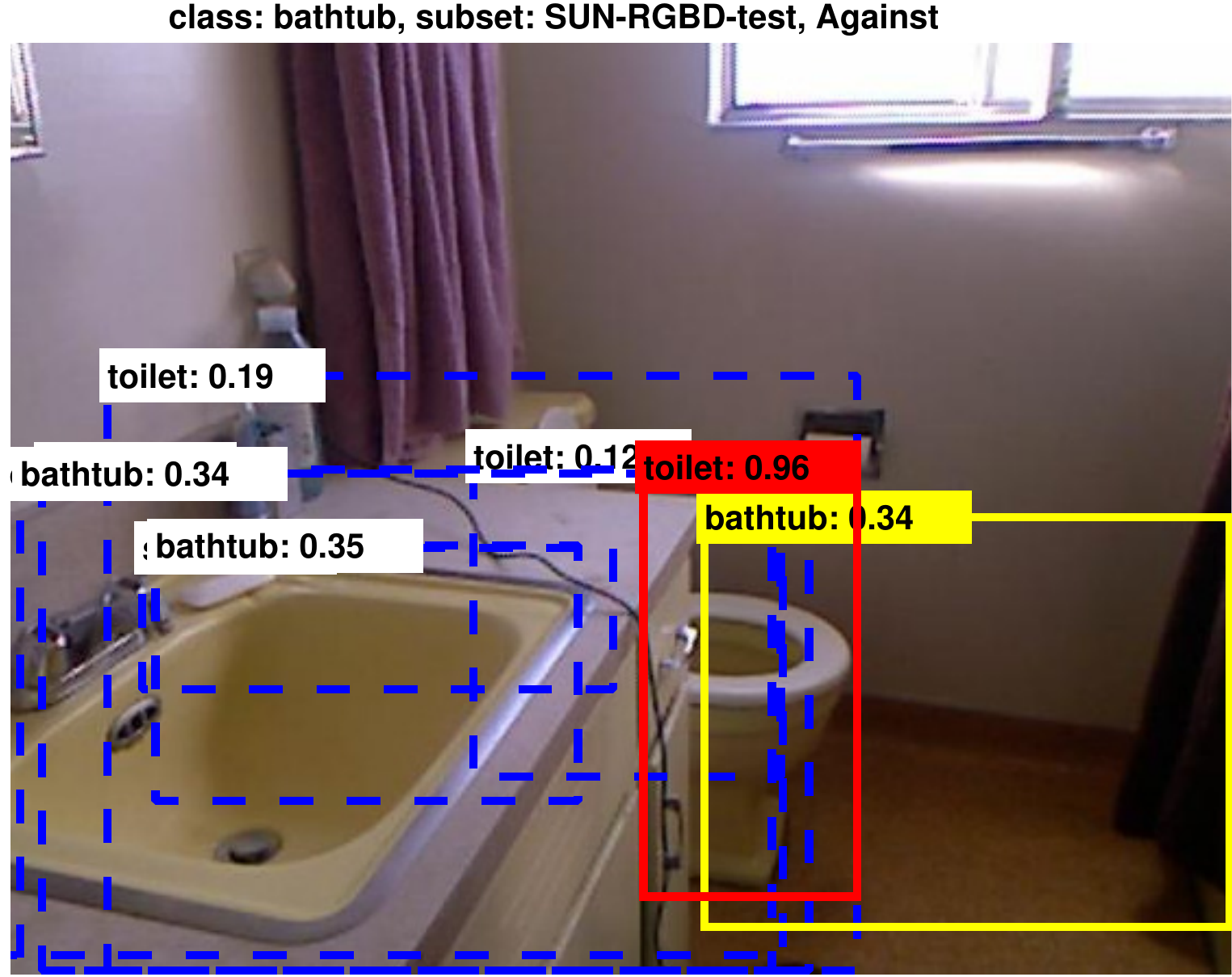}\label{fig:2748_bathtub_64}} 
    
    {\includegraphics[width=0.2\linewidth]{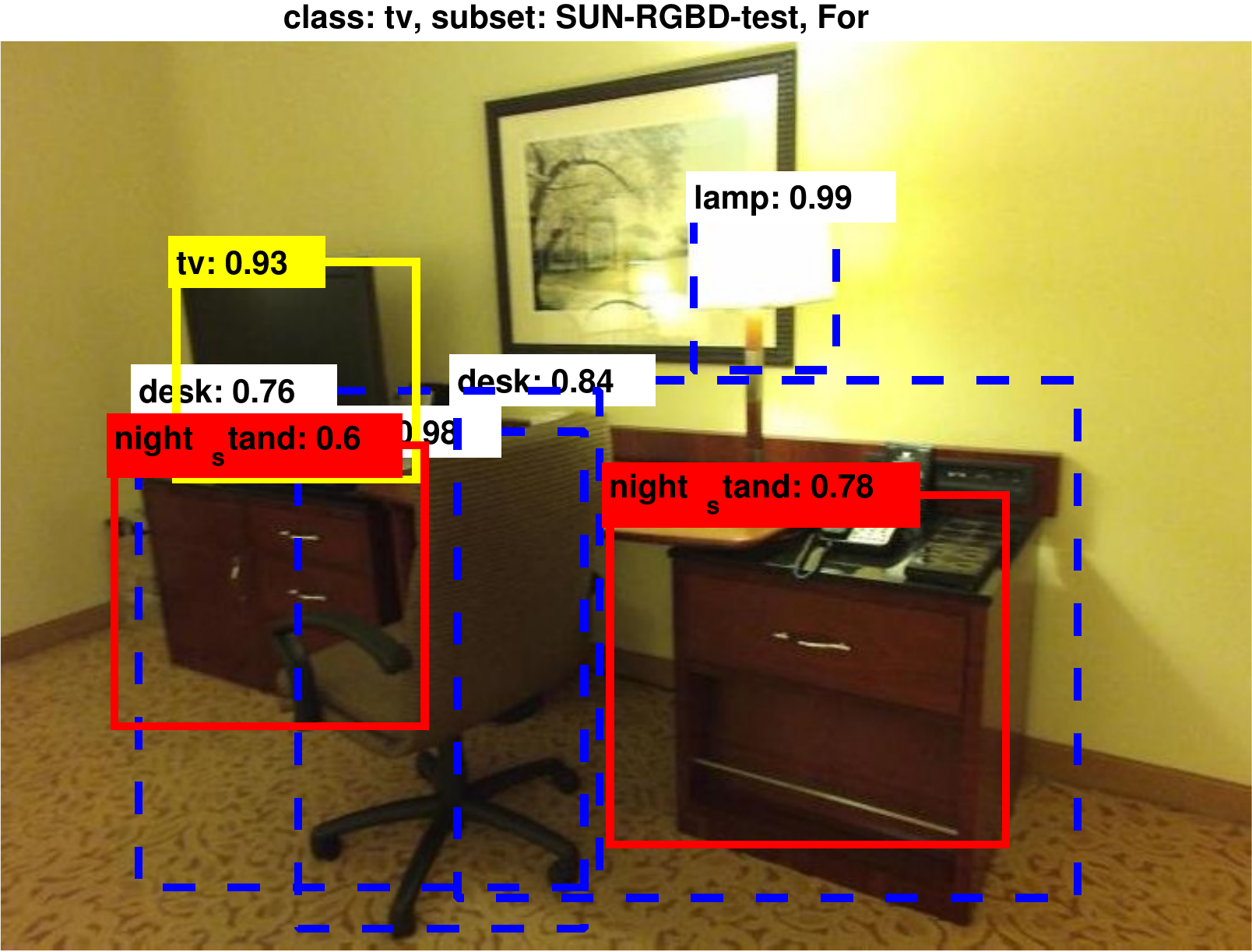}\label{fig:110_tv_16}}
    {\includegraphics[width=0.2\linewidth]{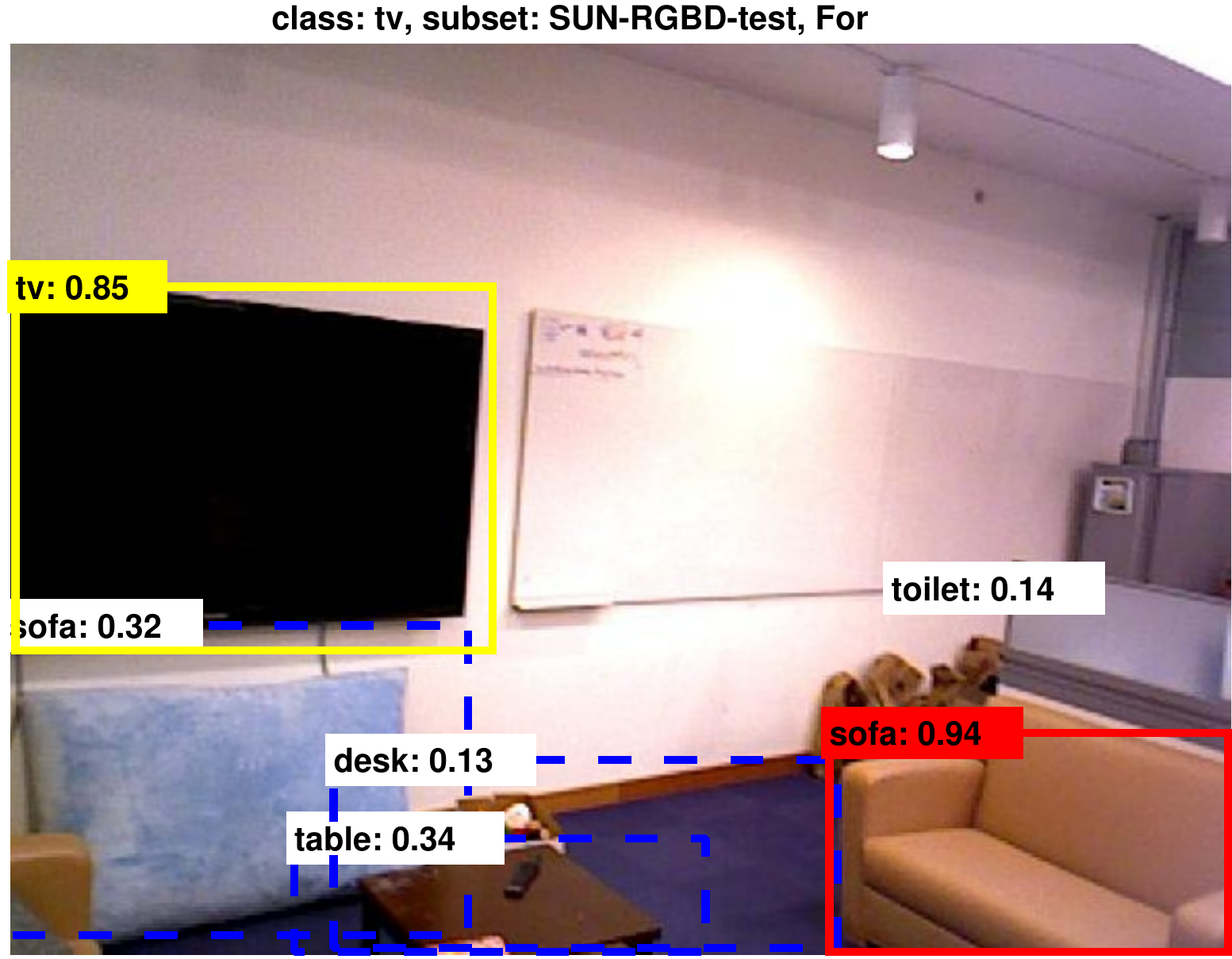}\label{fig:2983_tv_28}}
    {\includegraphics[width=0.2\linewidth]{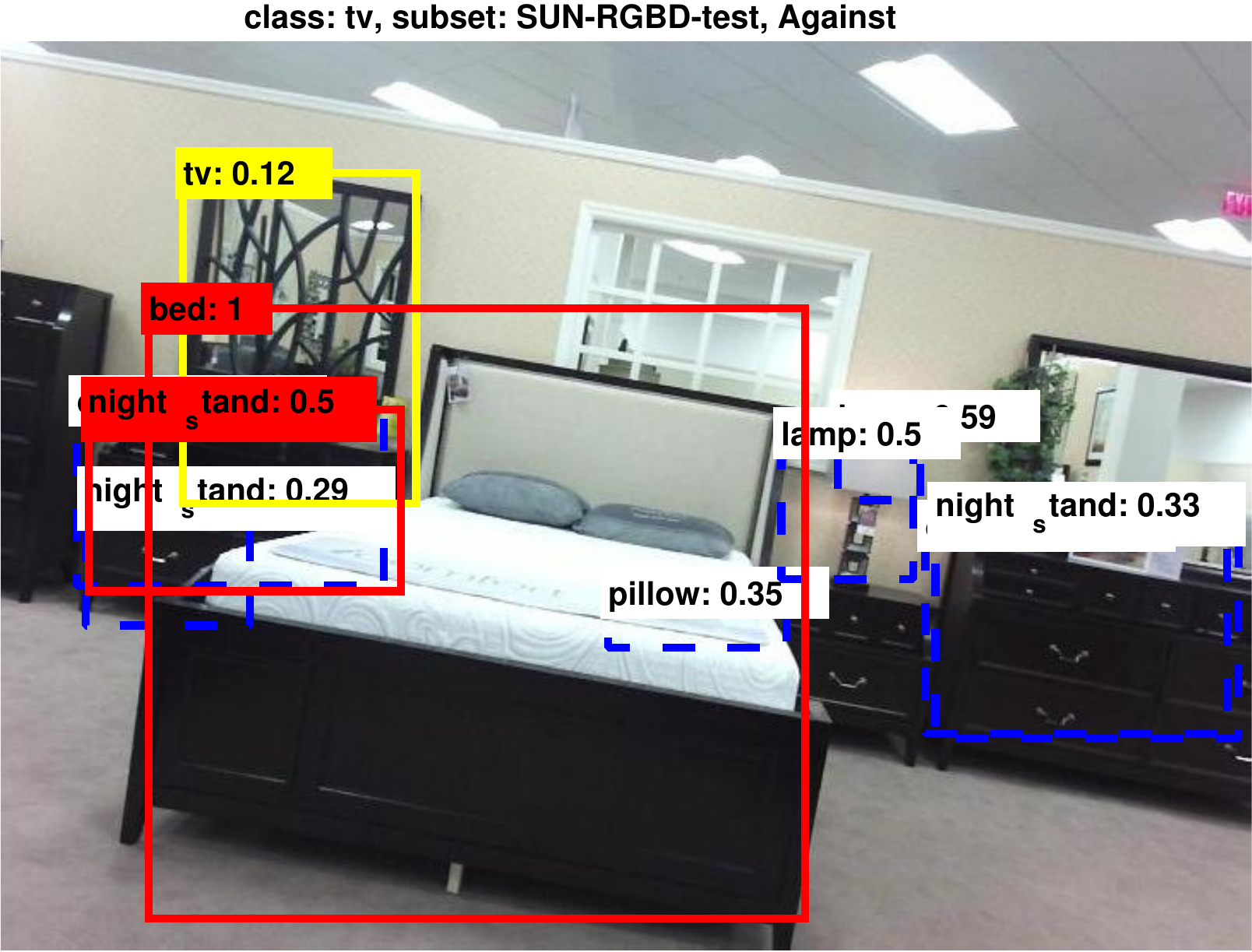}\label{fig:3_tv_27}}
    {\includegraphics[width=0.2\linewidth]{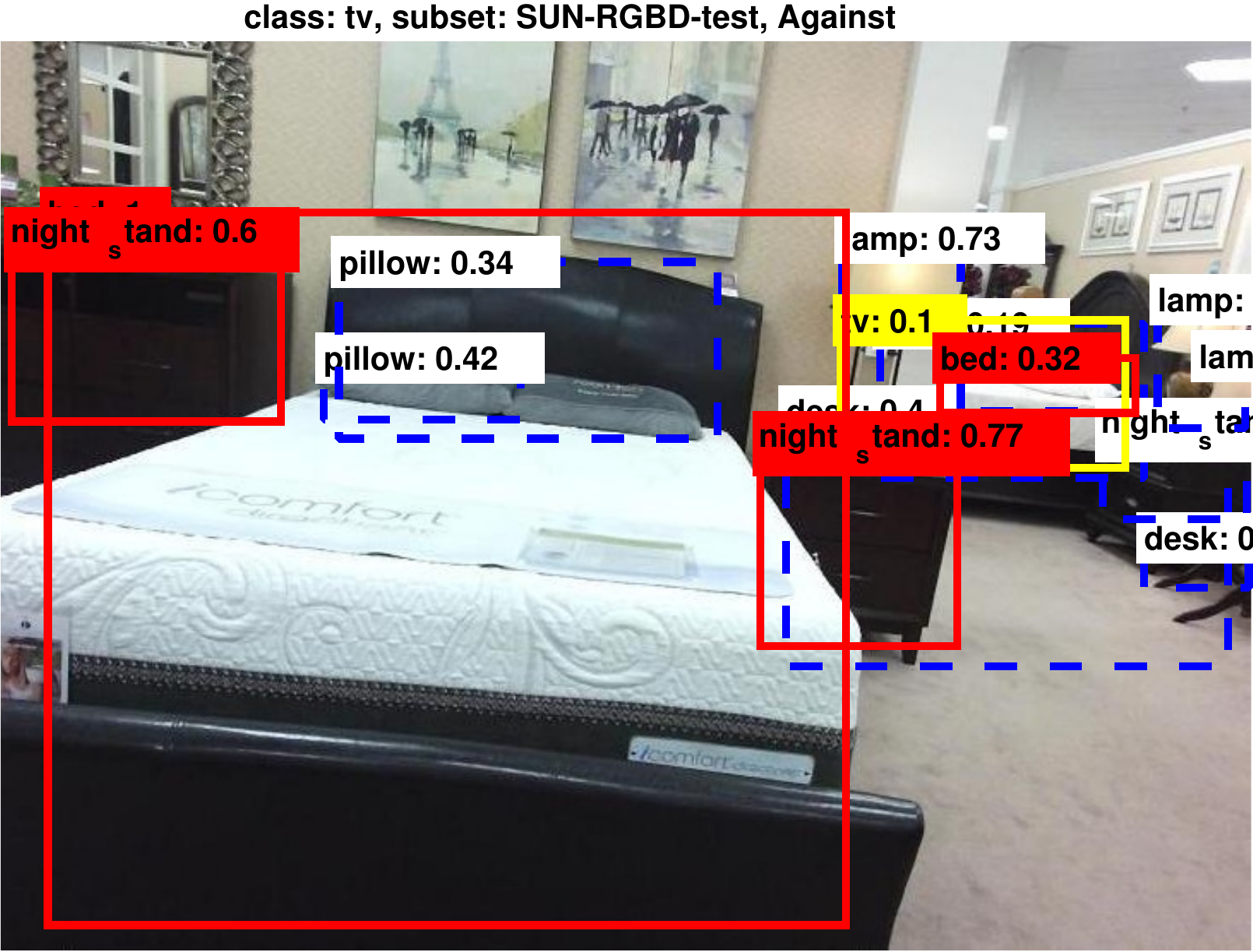}\label{fig:47_tv_48}}    
     \caption{\textbf{Visualization of For-Against Context Selection} We visualize the selected contextual regions with the labels that have the highest appearance-based confidence scores among all possible labels for four classes: \textit{bookshelf}, \textit{desk}, \textit{bathtub} and \textit{tv}. All figures are drawn when the precision threshold for potential contextual regions is set as 0.4. The first two columns show the selection results based on the FUB model and the last two columns are the results from the AUB model. The yellow boxes are the target objects, the red boxes are the selected contextual regions, and the blue dashed boxes are the ones that are not selected. } 
\label{fig:visl}
\end{figure}

\textbf{Visualization of Selected contextual regions} To visualize the performance of the context selection method, we show the selected contextual regions for four target object classes in Figure \ref{fig:visl}. The selection model tends to gather context information from the true positive contextual regions that can provide strong supporting or refuting evidence to predict the label of the target object. More visualizations can be found in the supplementary material.
\begin{table*}[h!]\tiny
  \centering
  \vspace*{-1mm}
  \caption{\textbf{Average Precision (AP) on SUN RGB-D Test Set: with Oracle.} CS is the full context model with selection that uses the margins between the two upper-bounds as confidence scores. SA-O is the select-all model with an oracle which labels true positive contextual regions. CS-O is the full context model with the oracle. The CS-O method outperforms the SA-O by allowing to select informative context regions, and its performance can be viewed as the upper-bound for the proposed CS model.}
  \label{tab:oracle} 
  \begin{tabular}{p{0.45cm}<{\centering} p{0.18cm}<{\centering} p{0.18cm}<{\centering} p{0.18cm}<{\centering} p{0.18cm}<{\centering}  p{0.18cm}<{\centering} p{0.18cm}<{\centering} p{0.18cm}<{\centering} p{0.18cm}<{\centering} p{0.18cm}<{\centering} p{0.18cm}<{\centering} p{0.18cm}<{\centering} p{0.18cm}<{\centering} p{0.18cm}<{\centering} p{0.18cm}<{\centering} p{0.18cm}<{\centering} p{0.18cm}<{\centering} p{0.18cm}<{\centering} p{0.18cm}<{\centering} p{0.18cm}<{\centering} p{0.18cm}<{\centering} }
   \hline
   
   & bathtub & bed & book-shelf & box & chair & counter & desk & door & dresser & garbage bin & lamp & monitor & night stand & pillow & sink & sofa & table & tv & toilet & mAP\\
   
 \hline
 CS   & 69.15 & 80.09 & 45.94 & 22.33 & 42.04 & 35.57 & 29.84 & 24.44 & 40.85 & 53.51 & 53.67 & 48.96 & 60.18 & 54.19 & 24.60 & 48.99 & 48.86 & \textbf{51.06} & 82.50 & 48.25 \\
   \hline  
   SA-O  &  70.07 & 78.95 & 45.48 & \textbf{23.29} & 42.91 & 36.97 & \textbf{33.59} & 24.37 & \textbf{42.39} & 52.44 & 52.78 & 49.15 & \textbf{61.81} & 53.38 & \textbf{26.49} & \textbf{53.44} & 48.71 & 49.16 & 81.73 & 48.79 \\
   \hline
   CS-O  &\textbf{73.67} & \textbf{80.56} & \textbf{50.42} & 23.08 & \textbf{47.51} & \textbf{37.33} & 30.98 & \textbf{24.98} & 41.03 & \textbf{54.45} & \textbf{56.49} & \textbf{50.08} & 60.25 & \textbf{55.29} & 25.30 & 49.09 & \textbf{49.27} & 45.91 & \textbf{83.55} & \textbf{49.43} \\
   \hline
  \end{tabular}
\end{table*}

 \vspace*{-2.5mm}
\begin{table}[htp]\scriptsize
\floatsetup{floatrowsep=qquad,captionskip=5pt} \tabcolsep=9pt
\begin{floatrow}
\ttabbox{\caption{Selecting Ratio: $T$ = Pillow}\label{select:pillow}}{%
\begin{tabular}{p{0.4cm}<{\centering} p{0.4cm}<{\centering} p{0.4cm}<{\centering} p{0.4cm}<{\centering} p{0.4cm}<{\centering} p{0.4cm}<{\centering}}\hline
   &  bed& 	sofa 	&pillow &	night-stand	&   lamp\\
   \hline
   Ratio & 0.92 & 0.89&0.84&0.79&	0.74  \\
 \hline
\end{tabular}}
\ttabbox{\caption{Selecting Ratio: $T$ = Bookshelf}\label{select:book}}{%
\begin{tabular}{p{0.4cm}<{\centering} p{0.4cm}<{\centering} p{0.4cm}<{\centering} p{0.4cm}<{\centering} p{0.4cm}<{\centering} p{0.4cm}<{\centering}}\hline
   & desk & table&book-shelf &chair& sofa  \\
   \hline
   Ratio & 0.97	&0.92 &0.81&0.77&	0.76  \\
 \hline
\end{tabular}}
\end{floatrow}
\end{table}

\section{Conclusion}
We analyzed the predictive potential of context in an idealized case where the labels of all contextual objects are known, and only these labels and their relationships to the target objects are used to predict the target object label. Through these experiments we found that, despite ignoring the appearance of the target object, pure context is effective at predicting the target object. We also discovered that different categories vary in their ability to predict a certain target object class. Based on these experiments, we proposed a region-based context re-scoring method with dynamic context selection to automatically improve the pool of contextual objects. Our method achieved significant performance gains when compared with the appearance-based detector and the contextual model that simply selects everything. An interesting direction of the future work is to use depth information as a contextual cue, and apply context selection in an end-to-end deep learning framework.

\section{Acknowledgement}
This research was partially supported by grant N000141612713 from the Office of Naval Research.

\bibliography{bmvc_final}
\end{document}